\pdfoutput=1
\documentclass[11pt]{article}

\usepackage[]{latex/acl}
\usepackage{times}
\usepackage{latexsym}

\usepackage[T1]{fontenc}
\usepackage[utf8]{inputenc}

\usepackage{microtype}

\usepackage{inconsolata}

\usepackage{graphicx}
\usepackage{booktabs}
\usepackage{array}
\usepackage{wrapfig}
\usepackage{multirow}
\usepackage{amssymb}
\usepackage{amsmath}

\usepackage{caption}
\usepackage{subcaption}
\usepackage{algorithm}
\usepackage{multirow}
\usepackage[noend]{algorithmic}
\usepackage{wrapfig}
\usepackage{framed}
\usepackage{pifont}

\newcommand{\intentsim}{\textsc{Intent-Sim}}

\ifx\nocomment\undefined
\NewDocumentCommand{\ec}
{ mO{} }{\textcolor{cyan}{\textsuperscript{\textit{Eunsol}}\textsf{\textbf{\small[#1]}}}}
\NewDocumentCommand{\mz}
{ mO{} }{\textcolor{blue}{\textsuperscript{\textit{Michael}}\textsf{\textbf{\small[#1]}}}}

\else
    \NewDocumentCommand{\ec}
    { mO{} }{\textcolor{cyan}{}}
      \NewDocumentCommand{\mz}
    { mO{} }{\textcolor{blue}{}}
\fi

\newcommand{\bettershortstack}[2][c]{%
  \begin{tabular}[b]{@{}#1@{}}#2\end{tabular}%
}
\title{Clarify When Necessary: \\ Resolving Ambiguity Through Interaction with LMs}

\author{Michael J.Q. Zhang \and Eunsol Choi \\
  Department of Computer Science\\
  The University of Texas at Austin\\
  \texttt{\{mjqzhang,eunsol\}@utexas.edu}}

\begin{document}
\maketitle

\begin{abstract}
Resolving ambiguities through interaction is a hallmark of natural language, and modeling this behavior is a core challenge in crafting AI assistants.
In this work, we study such behavior in LMs by proposing a task-agnostic framework for resolving ambiguity by asking users clarifying questions.
Our framework breaks down this objective into three subtasks: (1) determining \emph{when} clarification is needed, (2) determining \emph{what} clarifying question to ask, and (3) responding accurately with the new information gathered through clarification.
We evaluate systems across three NLP applications: question answering, machine translation and natural language inference.
For the first subtask, we present a novel uncertainty estimation approach, \intentsim, that determines the utility of querying for clarification by estimating the entropy over user intents.
Our method consistently outperforms existing uncertainty estimation approaches at identifying predictions that will benefit from clarification.
When only allowed to ask for clarification on 10\% of examples, our system is able to double the performance gains over randomly selecting examples to clarify.
Furthermore, we find that \intentsim~ is robust, demonstrating improvements across a wide range of NLP tasks and LMs. Together, our work lays foundation for studying clarifying interactions with LMs.

\end{abstract}

%%%%%%%%%%%%%%%%%%%%%%%%%%%%%%%%%%%%%%%%%%%%%%%%%%
% Intro Figure
%%%%%%%%%%%%%%%%%%%%%%%%%%%%%%%%%%%%%%%%%%%%%%%%%%
\begin{figure*}
\small
\centering

\includegraphics[width=16cm]{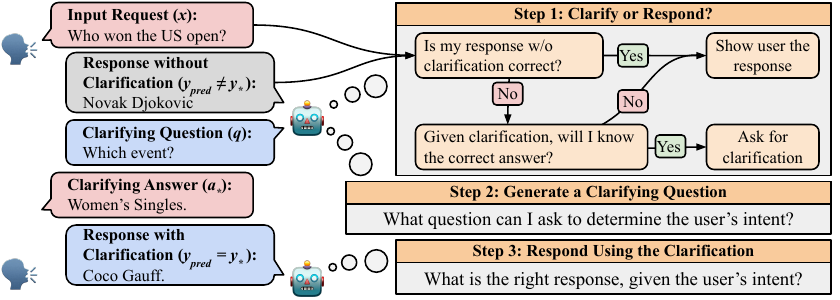}
% \vspace{-12pt}
\caption{
Our three-stage framework for resolving ambiguity with clarification questions.
In the first step, systems must identify which inputs will benefit from clarification.
In the second step, after deciding to clarify, we provide systems with a clarifying QA pair corresponding to the gold interpretation, which we generate from existing sources of disambiguated input/output pairs.
Finally, in the third step, systems use the input and the clarifying QA pair to arrive at the correct output.
}
\label{fig:intro}
% \vspace{-12pt}
\end{figure*}
%%%%%%%%%%%%%%%%%%%%%%%%%%%%%%%%%%%%%%%%%%%%%%%%%%

\section{Introduction}

Ambiguity is embedded throughout natural language, and even simple utterances can have multiple interpretations when read in isolation. Ambiguity serves a key, communicative function in language, allowing speakers to omit details by relying on information that is inferable from the extra-linguistic context of the conversation (e.g., temporal, social, and physical)~\cite{piantadosi2012communicative}. At times, however, the speaker's intent is still unclear despite the context. In such cases, further interaction is required to resolve the ambiguity, often by asking and answering clarifying questions.

With the recent progress in large language model (LLM) development, interactive AI assistants (e.g., ChatGPT, Claude, LLaMA-2) have risen to prominence in our daily lives; yet, these systems often fail to interact with users to resolve ambiguity in their requests. We address these shortcomings by establishing a task-agnostic framework for modeling and resolving ambiguity with LLMs using clarifying questions. We find that imbuing LLMs with the ability to ask clarifying questions can improve performance on a variety of NLP tasks.
Our framework breaks down the objective of resolving ambiguity into three sequential subtasks, which we depict in Figure~\ref{fig:intro}. In our first task, systems must decide \emph{when} to ask the user for clarification. We evaluate system for this task on their ability to maximize end-task performance while minimizing interaction cost. In our second task, systems must then decide \emph{what} to ask the users. Here, systems should ask questions that expose the ambiguity in the user's requests, eliciting a disambiguating response. Finally, after asking the user a clarifying question and receiving their response, systems perform the third and final task: producing the appropriate output given the ambiguous input and the user's clarification.

We apply this framework to a three of NLP settings: question answering (QA), machine translation (MT), and natural language inference (NLI). To cover each of these applications, we draw examples from existing datasets focused on modeling ambiguity~\cite{min2020ambigqa,bawden2018evaluating,liu2022were} and use multiple annotations to derive samples from the natural distribution over user intents for each ambiguous input. Having access to these natural distributions enables realistic evaluations, particularly for determining \emph{when} clarification is needed. Many ambiguous examples, while possessing multiple feasible interpretations, have only one mostly-likely interpretation that dominates the distribution over user intents (e.g., ``She's from Boston'' typically does not mean ``Boston, Georgia''). Systems for our first subtask should, therefore, be evaluated for their ability to identify such cases and avoid asking for unnecessary clarification.

We also develop systems for each of our subtasks, including an oracle method for generating clarification questions along with answers for different intents. We use this oracle to evaluate our other two subtasks, as performance on these tasks depends heavily on the quality of clarifying interactions.

Finally, we conclude our work by introducing \intentsim: a novel method for uncertainty estimation that we use to determine when to ask for clarification. \intentsim\ involves estimating the entropy over user intents by simulating multiple user-assistant interactions.
Through our experiments, we demonstrate the \intentsim\ consistently outperforms other uncertainty estimation baselines at identifying predictions that are both incorrect and can be improved with clarification.
We also find that these improvements are robust across different tasks and LLM systems.
\intentsim\ performs the best at identifying when asking for clarification will improve expected performance in four out of six of the LLM-plus-task settings evaluated in this work.
% \ec{add a link to a repo}

\section{A Framework for Resolving Ambiguity through Interaction}
We begin this work by formally defining three subtasks for resolving ambiguity with clarifying questions: (1) determining \emph{when} to ask for clarification, (2) identifying \emph{what} clarifying question to ask, and (3) reacting to clarification with the proper response. In Figure~\ref{fig:intro}, we depict how each of these three subtasks are sequentially applied in a user-assistant interaction to resolve ambiguity. This figure also depicts the notation used in this work, which we define below. 

% \vspace{-8pt}
\paragraph{Definitions}
Each interaction begins with the user providing an initial input request, $x$, to the LLM assistant.
Some inputs may be ambiguous, resulting in many feasible output responses for the system to choose from, which we denote as the set $Y = \{y_i\}_1^k$.
One of these outputs, $y_* \in Y$, represents the gold output corresponding to the user's intent behind their ambiguous request.
% Within this set of feasible outputs, one gold output, $y_* \in Y$, exists, corresponding to the user's intent behind their ambiguous request.
To determine the users intent, systems may ask the user a clarifying question, $q$.
The user then responds with the clarifying answer corresponding to their intent, $a_* \in A = \{a_i\}_1^k$.
For simplicity, we assume a bipartite matching between the sets of clarifying answers, $A$, and feasible final responses, $Y$.

Each input request $x$ has its own distribution over intended interpretations, $\mathbb{P}(y = y_* | x)$.
Accurately modeling this distribution is essential for avoiding asking unnecessary clarifying questions.
% Accurately modeling this distribution is essential for systems to avoid asking unnecessary and extraneous clarifying questions.
For example, when this distribution is dominated by a single feasible output, systems may want to forego clarification and respond to the user directly. Gathering annotations for the true distribution over intents, however, is intractable and temperamental (i.e., subject to changes over time, location, and individual preferences). Instead of assuming that we have access to this gold distribution, we say that our dataset consists of $(x, y_*)$ tuples, where intents and their respective outputs, $y_*$, are sampled from this distribution. We describe the data generation process for creating these samples in Section~\ref{sec:data}.

\subsection{Task 1: Determining when Clarification is Necessary}
The frequency with which systems should ask for clarification depends on the demands of the domain and preferences of the user. In high-stake settings, we may want systems to frequently ask for clarification. Likewise, for time-sensitive issues, we may want to minimize the number of interactions.
As such, we do not treat determining when to ask for clarification as a classification task; instead, we evaluate this challenge as an uncertainty estimation objective.
While standard uncertainty quantification only cares about estimating the performance of a given prediction, our task requires estimating how much performance would increase if provided clarification on the input.

This task requires systems to disentangle the two factors that contribute to model uncertainty: epistemic and aleatoric uncertainty~\citep{cole2023selectively}. Epistemic uncertainty refers to uncertainty that is due to a lack of knowledge. In the tasks we consider, this may occur in questions about an entities the LLM hasn't seen or words it hasn't observed the translation of. Aleatoric uncertainty, on the other hand, refers to uncertainty that is the result of some intrinsic randomness in the output. This randomness is often due to ambiguity, which we resolve through interaction. Systems for this task must identify instances with high aleatoric uncertainty, where the user's intent is ambiguous, and low epistemic uncertainty, where has the knowledge required to respond after clarification.

Concretely, systems for this task must predict a scalar uncertainty estimate, $u(x)$, for each input, $x$, that correlates with how much performance is expected to improve after clarification. Whether predictions improve with clarification is dependent on the performance on the other two subtasks (i.e. the quality of the clarifying interaction and the system's ability to use it to produce the correct output). We address these dependencies in our descriptions of the other two subtasks below.

% The system's performance after clarification is dependent on its performance in the other two steps of this pipeline. Below, we define the two metrics we use for evaluating a system's uncertainty estimates for this task.

% For each input $x$, models must predict an scalar uncertainty estimate for each input, $u(x)$, that correlate with how much performance improves with clarification. The system's performance after clarification is dependent on its performance in the other two steps of this pipeline. Below, we define the two metrics we use for evaluating a system's uncertainty estimates for this task.
% Below, we define the two metrics we use for evaluating systems for this task. Note that whether system's output will improve with clarification depends on the performance of the other two steps of the pipeline: asking and answering clarifying questions and responding to clarification. As such, when presenting our experiments and results for this task, we will do so after presenting 

% \vspace{-8pt}
\paragraph{Evaluation Metric: Performance Under a Fixed Interaction Budget}
To evaluate this task, we provide systems with an interaction budget, $b \in [0, 100]$, and allow systems to ask clarification questions on $b\%$ of input examples.
We use each system's uncertainty estimates, $u(x)$, to determine top $b\%$ of candidate examples to provide clarification for, then evaluate system performance under this interaction budget.
This metric is closely related to those used in selective prediction~\citep{ElYaniv2010OnTF}, a uncertainty estimation task where low-confidence predictions are either withheld or passed onto a human oracle to annotate by hand~\citep{tran2022plex}.
% \ec{maybe contextualize a bit with citation (selective QA, some others)?}

% \vspace{-8pt}
\paragraph{Evaluation Metric: AUROC}
Area under the receiver operating characteristic (AUROC) is a metric that commonly used in standard uncertainty quantification, where it is used to evaluate an uncertainty estimator's ability to classifying correct and incorrect predictions over all possible confidence thresholds. In our setting, we adapt this metric to evaluate the uncertainty estimate's ability to identify whether or not performance on an example will improve with clarification.

\subsection{Task 2: Generating Clarifying Questions and Answers}
%\ec{maybe we can motivate we don't focus on this as there have been prior work on this, while the other two tasks are novel and less studied?}
%\mz{that was the attempt... I tried to make it more clear!}

After determining whether or not clarification is needed, the next step is generating a clarifying question to ask the user and receiving their response. Numerous prior works have explored the task of generating a clarifying question based on an input, particularly in classification~\citep{Yu2019InteractiveCB}, FAQ~\citep{rao2018learning}, and moral assesment~\citep{pyatkin2022reinforced} domains. Given the depth of prior work on the subject, we do \emph{not} propose or evaluate new methods for generating clarifying questions conditioned on the input. Instead, we develop an oracle prompting method for generating clarifying questions and answers for different user intents, which we describe Section~\ref{sec:generating_clarifications}.
% \ec{should we lower expectation here? all we do is prompting..}\mz{Added "prompting"}

The purpose of developing this oracle is to establish a stable test-bed for evaluating systems on the other two subtasks. As mentioned above, the performance of each step in our clarifying pipeline is inextricably linked to the performance of the other two steps: accurately responding to clarifications requires high quality clarifying interactions, and determining when to ask for clarification also depends on the utility of the clarifying interactions. While prior work has established methods evaluating clarifying question generation, our other two subtasks are more novel and less well studied in LLMs. Therefore, our focus is on evaluating performance on our first and third tasks in isolation, using our oracle-generated clarifying interactions to limit the dependence on this intermediate step.

\subsection{Task 3: Responding to Clarifications}
In this final task, systems must use the input and the clarifying question and answer to arrive at the appropriate response. To evaluate this task, we simply evaluate the LLMs generated output $\hat{y}$, which is conditioned on the ambiguous input and the clarifying QA pair, against the gold output $y_*$. We use different metrics for comparing $\hat{y}$ against $y_*$ for each target task, which we describe in Section~\ref{sec:data}.
% We evaluate this task on all interpretations of all inputs, weighting each interpretation in two ways.

We evaluate systems under two data generation processes for sampling $(x, y_*, q, a_{*})$ examples. The first, \textsc{sampled}, is our standard setting, using samples from the true distribution of intended interpretations as described above. While this setup is well suited for estimating system performance in realistic settings, it can also underestimate the importance of achieving high performance at the tails of the distributions over intents. To avoid over-indexing on only the most common interpretations, which may lead to misleading and biased responses as a whole, we introduce the second setting, \textsc{uniform}, where we evaluate on all interpretations of each input, weighing each equally.

\begin{table*}
\centering
\footnotesize

\begin{tabular}{c c p{7.5cm}  c}
\toprule
Task & Ambiguity Type & Input (\textbf{\emph{x}}) and Clarifying Question (\textbf{\emph{q}}) & Proportion \\ \midrule
\multirow{9}{*}{QA}
& \multirow{3}{*}{\shortstack{Word-Sense \\ Disambiguation / \\ Entity Linking}}
& \textbf{\emph{x}:} Who wins at the end of friday night lights?
& \multirow{3}{*}{48\%} \\
& & \textbf{\emph{q}:} Are you referring to the Friday Night Lights film, book, or television series? & \\ \cmidrule{2-4}
& \multirow{3}{*}{\shortstack{Literal vs. Implied \\ Interpretation}}
& \textbf{\emph{x}:} Real name of gwen stacy in amazing spiderman?
& \multirow{3}{*}{8\%} \\
& & \textbf{\emph{q}:} Are you asking for the name of the actress who plays Gwen Stacy, or the full name of the character Gwen Stacy? & \\ \cmidrule{2-4}
& \multirow{2}{*}{\shortstack{Multiple Valid \\ Outputs}}
& \textbf{\emph{x}:} When did west germany win the world cup?
& \multirow{2}{*}{44\%} \\
& & \textbf{\emph{q}:} Which time? & \\ \midrule

\multirow{7.5}{*}{NLI}
& \multirow{4}{*}{\shortstack{Word-Sense \\ Disambiguation}}
& \textbf{\emph{x}:} Every night, the baby is fed milk. / Some nights, the baby is fed milk.
& \multirow{3}{*}{44\%} \\
& & \textbf{\emph{q}:} Does the baby get fed milk every night or just some nights? & \\ \cmidrule{2-4}
& \multirow{3}{*}{\shortstack{Literal vs. Implied \\ Interpretation}}
& \textbf{\emph{x}:} The cake was so dry, it was like eating sand. / The cake was so dry, it was inedible.
& \multirow{3}{*}{56\%} \\
& & \textbf{\emph{q}:} Was the cake not suitable for eating or not safe to eat? & \\ \midrule

\multirow{2}{*}{MT}
& \multirow{2}{*}{\shortstack{Word-Sense \\ Disambiguation}}
& \textbf{\emph{x}:} It's a little steeper than I was expecting.
& \multirow{2}{*}{100\%} \\
& & \textbf{\emph{q}:} What kind of mole are you referring to?  \\

\bottomrule

\end{tabular}
% \vspace{-0.25cm}
\caption{Example instances for each task for different ambiguity types, along with what proportion of ambiguities in each dataset fall into each type from our manual analysis on 150 examples.}
\label{tab:dataset_examples}
\vspace{-0.5cm}
\end{table*}
%%%%%%%%%%%%%%%%%%%%%%%%%%%%%%%%%%%%%%%%%%%%%%%%%%

\section{Datasets and Applications}\label{sec:data}
We apply our framework to three tasks and datasets for modeling ambiguity. All datasets label ambiguous inputs with their different interpretations, given as disambiguated rewrites or as different contexts, along with their respective outputs. We use these annotations later in developing our oracle system for generating clarifying questions and answers. All datasets lack existing labels for the distribution over these intents. Below, we describe each dataset in detail, as well as our methods for sampling intents for each example. We include dataset details in Appendix~\ref{app:data_details}.
% We apply our framework to three tasks and datasets for modeling ambiguity. All datasets label ambiguous inputs with their different interpretations, either as rewritten versions of the input or by providing different contexts for it, along with their respective outputs. We use these annotations later in developing our oracle system for generating clarifying questions and answers. Another commonality between these dataset is the lack of existing labels for the distribution over these intents. Below, we describe each dataset in detail, as well as how we sample of true intents for each example. We also include additional dataset details in Appendix~\ref{app:data_details}.

\subsection{Question Answering}
We use the AmbigQA~\citep{min2020ambigqa} dataset, which re-annotates questions from NaturalQuestions~\citep{kwiatkowski2019natural} with whether they are ambiguous. For each ambiguous example, they also annotate different intents as disambiguated revisions of the initial question, paired their respective answers. To draw from the true distribution over intents, we use the original annotated answers
% \ec{how many? if its dev set i assume 5..?}\mz{We take 1 sample ambiguous input, cleared this up in the appendix}
from NaturalQuestions as samples of intended outputs, $y_*$. We then map these sampled outputs to their respective intents by identifying which disambiguation contains the same answer.

% \vspace{-8pt}
\paragraph{QA Performance Metric}
We evaluate performance for QA using answer recall, measuring whether the gold answer string appears in the LLM's generated output after normalization~\citep{chen2017reading}. This deviates slightly from prior work~\cite{Rajpurkar2016SQuAD1Q} that evaluates for strict exact match after normalization, as chat-based LLMs to generate verbose, sentence-length outputs as opposed to short answers (e.g., ``The stern is the back of the boat.'' instead of ``the back''). 

\subsection{Natural Language Inference}
We source NLI data from the AmbiEnt dataset~\citep{liu2022were}, which consists of ambiguous premise/hypothesis pairs that are paired with disambiguated revisions for each of their feasible interpretations. Annotators for this dataset are first presented with the ambiguous input and are asked to label it as an NLI example. Annotators are then shown the different disambiguations each input, and are asked to label each interpretation again. We use these multiple annotations to identify which interpretation's label is consistent with the label annotators gave the initial, ambiguous input. We then use the matching interpretation as our sampled user intent and output label, $y_*$.
% We use this to identify which way the annotator interpreted each ambiguous input

% We source NLI data from the AmbiEnt dataset~\citep{liu2022were}, which consists of 
% ambiguous premise/hypothesis pairs that are paired with their disambiguated, rewritten interpretations and their respective outputs. Each ambiguous and rewritten premise/hypothesis pair is then 9-way annotated. As each output annotation $y_i$ can be mapped with interpretation $a_i$, we can get an estimated distribution over the user intents by mapping 9-way annotation.

% To estimate the distribution over interpretations, we compare the label each annotator assigned to the ambiguous input to the labels they assigned to each interpretation. We then determine that the annotator interpreted the input one way if they labeled the original input and an interpretation the same way. We then use this as the distribution over annotator readings of the input as the distribution over user intents. 
% \vspace{-8pt}
\paragraph{NLI Performance Metric} We evaluate systems using standard 3-way (entailment, contradiction, neutral) classification accuracy.

\subsection{Machine Translation}
The meaning of a sentence can be ambiguous when presented in isolation, but becomes clear in its document-level context.
Rich previous works have explored when sentence-level translation to fail in context~\citep{lopes2020document,yin2021does,voita2019good}.
We source examples of ambiguous translations from the DiscourseMT dataset~\citep{bawden2018evaluating}, a manually crafted test set of ambiguous English-French translations.
% We source examples of ambiguous translations from DiscourseMT dataset~\citep{bawden2018evaluating}, a manually crafted set of ambiguous English-French translations drawn from the OpenSubtitles dataset~\citep{lison-tiedemann-2016-opensubtitles2016}.
% For each context sentence, it contains a translation of the context and test sentence pair.
Each example consists of an ambiguous test sentence paired with two possible context sentences, where the translation of the test sentence depends on which context sentence precedes it.
We use these test sentences, without context, as examples of ambiguous user inputs, taking its two possible translations as the set of feasible outputs. We also include the context sentences, which are only annotated with one feasible translation each, as examples of unambiguous inputs. While this dataset does not contain annotations for estimating distribution over interpretations, sentences in this dataset are hand-crafted to be highly ambiguous. We, therefore, simply use the uniform distribution over interpretations in our experiments.

% We use the sentences with multiple translations examples of an ambiguous inputs, and each of its translations as its interpretations and respective outputs. We also include context sentences with only one translation as examples of unambiguous inputs. As this dataset does not contain annotations enabling us to estimate the distribution over interpretations, we simply use the uniform distribution over interpretations of ambiguous inputs in our experiments. %this examples in this dataset are designed to be highly ambiguous and 

% \ec{distribution simulation part?}% While this dataset does not contain annotations over the likelihood of each interpretation of the input, we, where each sentence's translation is minimally-edited based on the context
% \vspace{-8pt}
\paragraph{MT Performance Metric}
We evaluate using contrastive accuracy~\citep{maruf2019selective}. This binary metric measures whether an LLM assigns a greater likelihood to the intended translation of an ambiguous sentence over the alternative. For unambiguous examples, we simply say that the system gets the interpretation correct without clarification. We deviate from the standard MT metrics (e.g., BLEU), as confounding factors such as variance in sentence structure often overshadow the word-level, semantic differences between translations. 
%%%%%%%%%%%%%%%%%%%%%%%%%%%%%%%%%%%%%%%%%%%%%%%%%
% MT Followup Gen Prompt
%%%%%%%%%%%%%%%%%%%%%%%%%%%%%%%%%%%%%%%%%%%%%%%%%
\begin{table*}
\centering
\footnotesize

% \vspace{-0.1cm}
\begin{tabular}{c c l l l l l }
\toprule
\multirow{2.5}{*}{Model} & \multirow{2.5}{*}{Clarification} & \multicolumn{1}{c}{MT} & \multicolumn{2}{c}{QA} & \multicolumn{2}{c}{NLI} \\ \cmidrule(lr){3-3} \cmidrule(lr){4-5} \cmidrule(lr){6-7}
 &  & \multicolumn{1}{c}{Uniform} & \multicolumn{1}{c}{Uniform} & \multicolumn{1}{c}{Sampled} & \multicolumn{1}{c}{Uniform} & \multicolumn{1}{c}{Sampled} \\ \midrule
\multirow{3}{*}{GPT3}            & Direct   & 50.0  & 22.7  & 51.8  & 31.2  & 41.7  \\
                                 & Follow   & 85.8 (35.8) & 40.8 (18.1) & 61.8 (10.0) & 31.6 (0.4) & 45.9 (4.2) \\
                                 & Disambig & 84.7 (34.7) & 41.2 (18.5) & 62.0 (10.2) & 30.6 (-0.6) & 30.6 (-11.1) \\ \midrule
\multirow{3}{*}{LLAMA2 7B}       & Direct   & 50.0  & 14.5  & 31.4  & 29.4  & 32.4  \\
                                 & Follow   & 46.6 (-3.4) & 27.3 (12.8) & 45.4 (14.0) & 25.4 (-4) & 35.9 (3.5) \\
                                 & Disambig & 45.5 (-4.5) & 25.7 (11.2) & 41.1 (9.7) & 29.8 (0.4) & 29.8 (-2.6) \\ \midrule
\multirow{3}{*}{\bettershortstack{LLAMA2 7B \\ Chat}}  & Direct   & 50.0  & 18.1  & 37.3  & 41.0  & 43.5  \\
                                 & Follow   & 43.2 (-6.8) & 32.0 (13.9) & 47.9 (10.6) & 55.3 (14.3) & 52.5 (9.0) \\
                                 & Disambig & 44.9 (-5.1) & 26.5 (8.4) & 42.0 (4.7) & 40.0 (-1.0) & 40.0 (-3.5) \\ \midrule
\multirow{3}{*}{LLAMA2 13B}      & Direct   & 50.0  & 17.7  & 39.1  & 30.6  & 37.4  \\
                                 & Follow   & 46.6 (-3.4) & 34.1 (16.4) & 53.7 (14.6) & 34.6 (4.0) & 43.1 (5.7) \\
                                 & Disambig & 47.2 (-2.8) & 32.4 (14.7) & 50.8 (11.7) & 30.2 (-0.4) & 30.2 (-7.2) \\ \midrule
\multirow{3}{*}{\bettershortstack{LLAMA2 13B \\ Chat}} & Direct   & 50.0  & 17.9  & 40.0  & 28.0  & 40.7  \\
                                 & Follow   & 40.9 (-9.1) & 33.5 (15.6) & 50.9 (10.9) & 49.1 (21.1) & 52.5 (11.8) \\
                                 & Disambig & 42.6 (-7.4) & 28.5 (10.6) & 45.2 (5.2) & 26.6 (-1.4) & 26.6 (-14.1) \\

\bottomrule
\end{tabular}

\caption{
% \ec{i changed this a bit, TA}
Performance on responsiveness to clarification. We evaluate three settings: providing the clarifying QA pair (Follow), disambiguated input (Disambig), and baseline (Direct) without clarifying information. For QA and NLI, we evaluate under two different data generation processes, either uniformly weighing all interpretations or using our sampled interpretations. We evaluate MT using contrastive accuracy, QA using EM accuracy, and NLI using 3-way classification accuracy.}
\label{tab:responsiveness}
\vspace{-0.5cm}
\end{table*}
%%%%%%%%%%%%%%%%%%%%%%%%%%%%%%%%%%%%%%%%%%%%%%%%%

\section{An Oracle for Generating Clarifying Questions}\label{sec:generating_clarifications}
We begin our discussion of systems, experiments, and results by introducing our oracle method for generating clarifying questions, which we use to establish a test bed for evaluating system on the other two tasks our pipeline.
% We develop an oracle method for generating clarifying questions to establish a test bed for evaluating system on the other tasks within our pipeline.
Our oracle makes use of few-shot prompting with GPT-3.5~\citep{chatgpt}, providing systems with instructions and two hand-written exemplars to accomplish the following task: Given the ambiguous input, $x$, and its different interpretations, each corresponding to a different output $y \in \{y_i\}_1^k$, systems must generate (1) clarifying question differentiating each interpretation, $q$, and (2) then a clarifying response, $ \{a_i\}_1^k$, corresponding to each interpretation. The format of the different interpretations used as input to this system depend on the available annotations in each dataset: we use disambiguated revisions of $x$ for QA and NLI and the different target translations, $\{y_i\}_1^k$, for MT. This is an oracle setting, as it requires access to the different feasible interpretations of each input. We minimally edit the prompts between each tasks to reflect the different inputs and and interpretation formats from each task. See Appendix~\ref{app:model_details} for prompts and details.

% as annotated in the datasets described above (i.e., disambiguated versions of the question or premise-hypothesis pair, different translations of the input), model is prompted to first provide a clarifying question differentiating each interpretation, $q$, then a clarifying response, $ \{a_i\}_1^k$, corresponding to each interpretation. This is oracle setting as we provide $ \{y_i\}_1^k$, which we do not have access in realistic settings. We minimally edit the prompts between each tasks to reflect the different inputs and and interpretation formats from each task. See Appendix~\ref{app:model_details} for prompts and details.
% \vspace{-10pt}
\paragraph{Clarifying Interaction Analysis} 

In Table~\ref{tab:dataset_examples}, we identify the most common causes of ambiguity by analyzing clarifying questions. 
The most common cause across all tasks is word-sense disambiguation.
In QA, where named entities are more common, this also commonly surfaces as entity linking ambiguities.
The second cause is due to the literal and implied interpretations of each input. In QA, this usually occurs when a question literally means something different from what the user probably meant to ask. In NLI, we find that this frequently occurs due to figurative language, where it is unclear whether the sentence should be interpreted literally.
In MT, however, we find these ambiguities in the source sentence can usually be captured in its translation.
% One notable exception exception is when translating idioms~\citep{stowe2022impli}, but we do not include such special cases in our dataset.
The last common cause we find is ambiguity due to multiple valid outputs.
This cause only applies to QA where only reporting one answer may mislead users.
We do not find this type of ambiguity in MT, where multiple translations of any sentence is a given, nor in NLI, where classes are designed to be mutually exclusive.

\begin{table*}
\centering
\footnotesize
\setlength{\tabcolsep}{2pt}
\begin{tabular}{r l l l c}
\toprule
\multicolumn{2}{c}{Input with Sampled Clarification Question} & \multicolumn{2}{c}{Simulated User Answers} & Likelihood \\ \midrule
\multirow{2}{*}{\textbf{\emph{x\textsubscript{MT}}:}} & \multirow{2}{*}{There, on the trunk.}
  & The large storage box at the back of a car. & \multirow{3}{*}{\Bigg\}} & \multirow{3}{*}{60\%}\\
& & The large storage compartment of a car. & \\
  \textbf{\emph{q\textsubscript{greedy}}:}
& \multirow{2}{*}{\bettershortstack[l]{What type of trunk are you \\ referring to?}}
  & The back of a car. & \\
& & A large suitcase or box for storage. & \multirow{2}{*}{\Big\}}& \multirow{2}{*}{40\%}\\
& & The large, wooden storage chest. & \\ \midrule
\textbf{\emph{x\textsubscript{QA}}:} & \multirow{2}{*}{\bettershortstack[l]{How many Grammy Awards \\ does Whitney Houston have?}}
  & \multirow{3}{*}{\bettershortstack[l]{The number of Grammy Awards \\ Whitney Houston won. \emph{\textbf{(Repeated $\times$ 4)}}}} & \multirow{3}{*}{\Bigg\}} & \multirow{3}{*}{80\%} \\
& &  & \\
  \textbf{\emph{q\textsubscript{greedy}}:}
& \multirow{4}{*}{\bettershortstack[l]{Are you referring to the number of \\ Grammy Awards Whitney Houston \\ won, or the number of Grammy Awards~~ \\ Whitney Houston was nominated for?
}} & \\ 
 & & \multirow{3}{*}{\bettershortstack[l]{The number of Grammy Awards \\ Whitney Houston was nominated for.}} & \multirow{3}{*}{\Bigg\}} & \multirow{3}{*}{20\%}\\ \\ \\
\bottomrule
\end{tabular}
\caption{Generations from our \intentsim~method. Systems greedily generate a clarifying question based on the input, then sample multiple user responses. We group equivalent responses using an NLI system, then compute the likelihoods and entropy over the grouped, simulated intents.}
\label{tab:user_sim}
% \vspace{-0.1cm}
\vspace{-0.5cm}
\end{table*}
% %%%%%%%%%%%%%%%%%%%%%%%%%%%%%%%%%%%%%%%%%%%%%%%%%

\section{Experiments: Responsiveness to Clarification}\label{sec:experiments}

\paragraph{Setting} We evaluate LLM variants for their responsiveness to clarifications by comparing their performance on ambiguous examples with and without clarification. We use standard few-shot prompting for all systems (LLaMA-2, LLaMA-2-Chat, GPT-3), providing them with demonstrations from each task with and without the clarifying QA pairs. We use four randomly sampled exemplars for each example and perform greedy decoding. The exact prompts are available in Appendix~\ref{app:model_details}.

% For chat-based LLM's, we split the same prompt into multiple messages to simulate a conversation history between a user and AI assistant.\ec{what do you mean here? unclear to me} Further details are available in Appendix~\ref{app:model_details}.

\paragraph{Results}
We report our results in Table~\ref{tab:responsiveness}. We find that, across tasks and systems, LLMs can leverage clarifying questions and answers to improve their response. One exception to this trend, however, is the performance of LLaMA-2 variants on MT. We attribute this poor performance after clarification to LLaMA-2's low translation performance due to insufficient multilingual pre-training.
% \ec{did we evaluate bleu scores? maybe report those?}% \mz{cite something here?}

Another notable trend is that systems tend to perform better with clarifying questions and answers than with disambiguated inputs, particularly for QA and NLI. We attribute this the way our QA and NLI datasets construct disambiguated interpretations. These datasets create disambiguated revisions of each ambiguous input by applying a minimal set of token-level edits to the initial input. While this makes disambiguations easier to annotate and compare, it comes at the cost of naturalness of the resulting disambiguated sentences. In contrast, our clarifying interactions do not have the same minimal-edit constraints and more closely resemble the pretraining distributions of these systems.

We also find that there is no consistent improvement in LLaMA-2 and LLaMA-2-Chat's ability to use clarifying interactions. While task performance changes as a whole with chat-finetuning, the gains from providing clarifications remains consistent between equal size LLaMA-2 systems. These findings reinforce our motivations for studying this problem, as existing systems struggle to use clarifying questions and existing methods for chat-finetuning do not adequately train this ability.

\section{Experiments: Determining When to Clarify}

% \ec{motivate our method. maybe say something like existing methods for uncertainty estimation does not consider how useful would the answers to the clarification would be like? etc }

For our experiments on determining when to clarify, we use the same base LLMs as above for answering questions with and without clarification.
We adapt existing methods for uncertainty estimation and chain-of-thought reasoning as baselines for this task.
We begin this section by describing our novel approach to this subtask, before introducing baselines below.

\subsection{\intentsim}

Unsupervised methods for uncertainty quantification in LLMs generally rely on estimating entropy over the output distribution, using high entropy to identify erroneous outputs~\citep{kadavath2022language,kuhn2023semantic}. While these methods perform well at identifying incorrect predictions, they fail to identify \emph{why} predictions are incorrect. Determining when to ask for clarification requires moving beyond simply identifying incorrect outputs and requires systems to attribute when uncertainty is the result of ambiguity. In our proposed method, \intentsim, we disentangle these two factors by explicitly estimating the ambiguity of a given input, which we quantify as the entropy over simulated user intents.

Figure~\ref{alg:user_sim} illustrates our method. Using the same few-shot prompt structure as in our responsiveness task (exact prompt in Appendix~\ref{app:model_details}), we condition on the user's request to greedily generate a clarifying question. We then simulate different user intents by sampling multiple responses to the clarifying question (example generations in Table~\ref{tab:user_sim}). Following \citet{kuhn2023semantic}, we then cluster sets of semantically equivalent responses using a DeBERTa-large NLI model~\citep{he2021deberta} finetuned on MNLI~\cite{multinli}. We say that two responses are equivalent if either clarifying QA pair entails each other, then estimate the likelihood of each set as the proportion of samples in it. Finally, we compute our uncertainty estimate by computing the entropy of this distribution over semantically distinct answers. In our experiments we decode 10 user responses with temperature $T=0.5$ for all systems, following follow prior work on estimating uncertainty in LLMs from samples~\citep{kuhn2023semantic, cole2023selectively}. Additional implementation details are provided in Appendix~\ref{app:model_details}.

\begin{figure}
% \vspace{-0.25cm}
\begin{framed}
\footnotesize
\vspace{-0.25cm}
\textbf{Input:} LM $M$, NLI model $N$, User input $\mathbf{x}$, sampling temperature $T$, and simulation count $S$. \\
\textbf{Output:} Entropy over simulated intents, $u$.

\begin{algorithmic}[1]
\STATE $\mathbf{q} \gets \textbf{GreedySample}(M, \left[ \mathbf{x} \right])$\label{alg:gen}
\FOR{$i  \in \{1, \dots, S \}$}
    \STATE $\mathbf{a_{i}} \gets \textbf{TempSample}(M, \left[ \mathbf{x}; \mathbf{q} \right], T)$
\ENDFOR
\STATE $G \gets \emptyset$
\FOR{$i  \in \{1, \dots, S-1 \}$}
    \FOR{$j  \in \{i+1, \dots, S \}$} 
        \STATE $\texttt{left} \gets N(\left[\mathbf{q}; \mathbf{a_i} \right], \left[\mathbf{q}; \mathbf{a_j} \right])$
        \STATE $\texttt{right} \gets N(\left[\mathbf{q}; \mathbf{a_j} \right], \left[\mathbf{q}; \mathbf{a_i} \right])$
        \IF{\texttt{left} is \texttt{entailment} or \texttt{right} is \texttt{entailment}}
            \STATE $G \gets G \cup \{<i, j>, <j, i>\}$
        \ENDIF
    \ENDFOR
\ENDFOR
\STATE $C \gets \emptyset$
\FOR{$i  \in \{1, \dots, S \}$}
    \IF{$\mathbf{a_{i}} \not\in c \ \ \ \forall c \in C$}
        \STATE $C \gets C \cup \textbf{DFS}(G, a_i)$
    \ENDIF
\ENDFOR
\STATE $\widehat P(c | \mathbf{x}) \gets \frac{|c|}{S}, \ \ \forall c \in C$
\STATE $u \gets \textbf{Entropy}(\widehat P(\cdot | \mathbf{x}))$
\end{algorithmic}
\vspace{-0.25cm}
\end{framed}
\vspace{-0.2cm}
% \caption{\intentsim~Psuedocode. We construct a graph of equivalent simulated responses, and identify the disjoint subgraphs as distinct intents.}
\caption{\intentsim~algorithm. We sample a clarifying question and responses from the LLM. We then construct a equivalence graph of responses, $G$, using NLI to determine equivalence. Finally, we identify disjoint subgraphs of $G$ with depth-first-search, representing distinct intents, and estimate the entropy over intents.
% \ec{DFS should be explained?}
}
\label{alg:user_sim}
\vspace{-0.5cm}
\end{figure}

%%%%%%%%%%%%%%%%%%%%%%%%%%%%%%%%%%%%%%%%%%%%%%%%%%
% WhenToAsk Results
%%%%%%%%%%%%%%%%%%%%%%%%%%%%%%%%%%%%%%%%%%%%%%%%%%
\begin{table*}
\centering
\footnotesize
% \vspace{-0.15cm}
\begin{tabular}{cclrrrrr}
\toprule
Task & Model & Method     & AUROC & $b=10\%$    & $b=20\%$    & $b=30\%$  \\ \midrule
\multirow{4}{*}{MT} & \multirow{4}{*}{GPT-3}
    & Likelihood & \textbf{0.547} & 76.1 (6\%) & 78.1 (17\%) & 79.8 (27\%) \\
 &  & Self-Ask & 0.371 & \textbf{77.3 (13\%)} & \textbf{79.5 (25\%)} & \textbf{81.5 (37\%)} \\
 &  & Sem. Ent & 0.531 & 76.4 (11\%) & 78.4 (19\%) & 80.4 (30\%) \\
 &  & User Sim & 0.512 & \textbf{77.3 (13\%)} & 78.7 (21\%) & 79.3 (24\%) \\ \midrule
\multirow{8.5}{*}{NLI} & \multirow{4}{*}{LLaMA-2 7B Chat} 
    & Likelihood & 0.416 & 41.2 (1\%) & 40.0 (-7\%) & 39.4 (-11\%) \\
 &  & Self-Ask & 0.477 & 41.6 (4\%) & 41.9 (7\%) & 42.5 (11\%) \\
 &  & Sem. Ent & 0.467 & 43.9 (21\%) & 44.1 (22\%) & \textbf{43.7 (19\%)} \\
 &  & User Sim & \textbf{0.531} & \textbf{44.3 (24\%)} & \textbf{44.3 (24\%)} & 43.1 (15\%) \\ \cmidrule{2-7}
 & \multirow{4}{*}{LLaMA-2 13B Chat} 
    & Likelihood & 0.526 & 31.0 (14\%) & 33.0 (24\%) & 33.8 (27\%) \\
 &  & Self-Ask & 0.462 & 28.2 (1\%) & 30.6 (12\%) & 34.0 (28\%) \\
 &  & Sem. Ent & 0.525 & 29.8 (8\%) &  \textbf{33.0 (24\%)} & 33.8 (27\%) \\
 &  & User Sim & \textbf{0.544} & \textbf{31.0 (14\%)} & 32.8 (23\%) & \textbf{34.8 (32\%)} \\ \midrule
\multirow{13}{*}{QA} & \multirow{4}{*}{GPT-3} 
    & Likelihood & 0.590 & 55.4 (14\%) & 55.9 (25\%) & 56.3 (35\%) \\
 &  & Self-Ask & 0.538 & 55.1 (6\%) & 55.6 (18\%) & 56.2 (32\%) \\
 &  & Sem. Ent & 0.625 & \textbf{55.5 (17\%)} & \textbf{56.1 (29\%)} & \textbf{57.0 (49\%)} \\
 &  & User Sim & \textbf{0.628} & \textbf{55.5 (17\%)} & \textbf{56.1 (29\%)} & \textbf{57.0 (49\%)} \\ \cmidrule{2-7}
 & \multirow{4}{*}{LLaMA-2 7B Chat} 
    & Likelihood & 0.510 & 38.4 (-1\%) & 39.1 (14\%) & 39.7 (28\%) \\
 &  & Self-Ask & 0.510 & 38.9 (10\%) & 39.3 (17\%) & 39.9 (32\%) \\
 &  & Sem. Ent & \textbf{0.532} & \textbf{39.1 (13\%)} & \textbf{39.3 (19\%)} & \textbf{40.1 (36\%)} \\
 &  & User Sim & 0.501 & 38.7 (6\%) & \textbf{39.3 (19\%)} & 39.7 (26\%) \\ \cmidrule{2-7}
 &  \multirow{4}{*}{LLaMA-2 13B Chat} 
    & Likelihood & 0.551 & 41.1 (8\%) & \textbf{41.7 (21\%)} & 41.8 (24\%) \\
 &  & Self-Ask & 0.546 & 41.0 (6\%) & 41.6 (20\%) & 42.1 (30\%) \\
 &  & Sem. Ent & 0.552 & 41.0 (6\%) & 41.4 (14\%) & 42.0 (28\%) \\
 &  & User Sim & \textbf{0.570} & \textbf{41.3 (11\%)} & 41.5 (17\%) & \textbf{42.8 (37\%)} \\
\bottomrule
\end{tabular}
\caption{Results for determining when to clarify. We report AUROC and performance under fixed interaction budget, $b$, evaluated using contrastive accuracy for MT, accuracy for QA and NLI. We also report the percent gain in performance relative to the total gain from clarifying all examples. }
\label{tab:when_to_ask_results}
\vspace{-0.5cm}
\end{table*}
%%%%%%%%%%%%%%%%%%%%%%%%%%%%%%%%%%%%%%%%%%%%%%%%%%

\subsection{Baselines}

\paragraph{Likelihood}
For this baseline, we first prompt the model to generate the answer without clarification using the same few-shot prompt as above. We then use the likelihood of the greedily generated output to determine when to clarify. This simple yet effective baseline is often used for uncertainty estimation, determining which model outputs are likely incorrect. In this work, we use low-certainty in the output as an indicator that clarification may improve the model's response.

\paragraph{Self-Ask}
Introduced by \citet{press2022measuring}, this prompting method is designed elicit chain-of-thought reasoning from LLMs for compositional reasoning tasks such as multi-hop QA. In their method, LLMs decompose inputs into multiple sub-questions and answers, which are composed to get the final answer. Self-Ask revolves around an intermediate step, where models decide whether to continue generating more questions or to complete their final response. We adapt this technique for our task, where the focus is not on decomposing the input but on querying for outside context. We adjust our few-shot prompt from our responsiveness task above and prompt assistants after each input query with the question ``Is a follow-up question needed here?'' (exact prompt in Appendix~\ref{app:model_details}). We then use the likelihood of generating ``No'' to score whether that clarification is needed. We also include this step in our sampled few-shot exemplars, creating a 50-50 split between unambiguous inputs, where the system responds ``No'', and ambiguous inputs, where systems respond ``Yes''.

\paragraph{Semantic Entropy}
We adapt this method from \citet{kuhn2023semantic}, which estimates the entropy over an LLM's output space by sampling multiple outputs and grouping equivalent responses. This method is closely related to our proposed \textsc{Intent-Sim} approach, only that we estimate entropy over the output space as opposed to over user clarifications. Following \citet{kuhn2023semantic}, we few-shot prompting the LLM to provide an output without clarifications, and determine equivalence between different sampled QA outputs by determining whether the QA pairs entail each other using the same pretrained NLI model as above. To apply this method to MT and NLI, we determine equivalence using exact match over sampled outputs. Then, following algorithm depicted in Figure~\ref{alg:user_sim}, we cluster outputs and compute entropy over equivalent output sets.

\subsection{Results}
In Table~\ref{tab:when_to_ask_results}, we report our results using our various LLMs and methods for deciding when to clarify. In comparing these systems gainst the random baseline,  which randomly selects $b\%$ of examples to clarify and achieves percent gain in performance equal to the budget value $b$, we observe that that Likelihood and Self-Ask demonstrate mixed results. While these systems generally perform better than random, the perform considerably worse under some budget settings. In contrast, using semantic entropy and simulating user interactions consistently outperforms all baselines, and outperform the random baseline under all interaction budgets.

Despite strong performances, we observe that this strong performance does not always translate similar gains in our AUROC metric. We attribute this gap to the coarse-grained nature of estimating entropy from samples. With only 10 samples, many examples produce the same distribution over clarifying answers equivalency sets. This is particularly true for examples where the entropy over clarifying answers is low. As a result, under very large values for $b$, entropy over simulated user responses may under perform compared to these other baselines; however, these larger values of $b$ are also less practical, as we aim for systems that ask questions conservatively, and $<50\%$ of examples in both datasets benefit from clarification.

\section{Related Work}

% \paragraph{Ambiguity and Uncertainty}
% Ambiguity in NLP has been explored in numerous prior works. Outside of the ones used in this work for generating datasets, several works have looked also at modeling annotator uncertainty in NLI~\cite{pavlick2019inherent,ynie2020chaosnli}. 

\paragraph{Clarifying Questions}
Selecting clarification questions has been previously studied in task-specific settings. \citet{rao2018learning} explores re-ranking clarification questions in product FAQ's, and \citet{shridhar-etal-2023-distilling} studies generating clarifying questions as a supervised learning task, using generated questions for multi-step reasoning in knowledge distillation. \cite{pyatkin2022reinforced} uses reinforcement learning (RL) to guide their question generation model toward proposing questions that can have a large effect on its moral judgment of a situation. Prior work~\citep{Yu2019InteractiveCB} studies balancing asking clarification questions and making the final classification prediction over multiturn interactions. 
Their clarifying questions only cover existing attributes, while ours are open ended.% We take a simpler approach to uncertainty estimation, studying a wide range of tasks using supervised data.
% \cite{press2022measuring} studies prompting LLMs to ask questions to perform chain-of-thought reasoning. Their focus, however, is primarily on generating questions to decompose the input into intermediate steps, and not to query for additional context.  

% \ec{discuss self ask  -- their is asking subquestion while we ask external information beyond what's asked in the question already}
% \vspace{-10pt}
\paragraph{Uncertainty Estimation}
Several existing works have studied methods for disentangling different sources of uncertainty. \cite{kamath2020selective} studies predicting out-of-distribution test examples, a source of epistemic uncertainty, and performing selective prediction by abstaining from predicting on such inputs. \cite{kuhn2023semantic} attempts to merge the likelihoods semantically equivalent in QA, eliminating the effect of uncertainty due to multiple vocalizations of the same answer. \cite{cole2023selectively} studies a similar setting in the intersection of ambiguity and uncertainty; however, this work does not consider the degree of ambiguity of various inputs, and does not attempt to resolve ambiguity through interaction. Other works have studied uncertainty estimation techniques for LLMs~\citep{kadavath2022language,lin2022teaching}, but they do not explicitly model or evaluate their ability to disentangle different sources of uncertainty. These works also explore supervised methods for uncertainty estimation in LLMs, but find that these methods generalize poorly to new domains.

% \vspace{-10pt}
\paragraph{Ambiguity in NLP}
Numerous prior works have created datasets for studying ambiguity in NLP, including work in coreference resolution~\citep{yuan2023ambicoref}, NLI~\citep{pavlick2019inherent}, and MT~\citep{pilault2023interactive}. The last work on MT also studies resolving ambiguity in an interactive chain-of-thought setting; however, it does not consider the challenge of modeling how ambiguous a given input is or determining whether interaction is helpful. Ambiguity benchmarks can also provide a lens to study biases. \cite{parrish2021bbq} studies an ambiguous QA task where systems are evaluated whether they resolve ambiguity by relying on harmful social biases.
% While we do not address this bias benchmark in this work, specifically identifying ambiguities and refraining from answering without proper context can be directly applied to mitigating bias as well.
% \mz{supervised methods have negative results}

% \ec{discuss datasets we considered but didn't use at the end? coref one, there was MT one too?}

% \paragraph{}

% Asking clarification question has been studied as a supervised learning task
% \cite{shridhar-etal-2023-distilling} introduces asking clarification questions focusing on multistep reasoning task to improve distilling from teacher model.  

% self-ask ofir stuff

% \paragraph{}

% ~\cite{rao}
% \cite{White2021OpendomainCQ}
% They introduce  asking clarification questions but for the purpose of classification task
% dialog inpainting 

% abstaining from answering, selective QA stuff

% \paragraph{Clarification Questions}

\section{Conclusion}
We present a unified framework for resolving ambiguity with clarifying questions, and apply it to QA, MT, and NLI. Our framework exposes the challenges in modeling clarifying interactions, and motivates the further study of disentangling uncertainty estimation and identifying when uncertainty can be attributed to ambiguity. We present a novel uncertainty estimation approach for this objective, \intentsim, which we demonstrate improves detection of when to clarify. 

Our framework lays the foundation for future work to explore interactive ambiguity resolution in general-purpose AI assistants. Future works using our framework may include developing new task-agnostic methods for generating clarification questions, or extending our framework to handle multi-turn interactions. Our work also motivates a closer examination of what is being learned through chat-finetuining for LLMs. Future work may develop new, multi-turn learning objectives using our framework and teach models to use interaction pragmatically, asking users clarifying question to maximize the accuracy of their responses.%In particular, our evaluations highlight shortcomings in these systems and the inability to resolve ambiguity when needed.

% Our framework focuses on evaluating two facets of interactive ambiguity resolution: determining when to ask for clarification and responding appropriately given clarification. 

% \ec{add future work}

\section*{Acknowledgements}
We thank the members of UT NLP community for early feedback on the draft, especially Jessy Junyi Li. This work was in part supported by Cisco Research. Any opinions, findings and conclusions, or recommendations expressed in this material are those of the authors and do not necessarily reflect the views of Cisco Research.

\bibliography{anthology,custom}

\begin{thebibliography}{30}
\expandafter\ifx\csname natexlab\endcsname\relax\def\natexlab#1{#1}\fi

\bibitem[{Bawden et~al.(2018)Bawden, Sennrich, Birch, and Haddow}]{bawden2018evaluating}
Rachel Bawden, Rico Sennrich, Alexandra Birch, and Barry Haddow. 2018.
\newblock Evaluating discourse phenomena in neural machine translation.
\newblock In \emph{16th Annual Conference of the North American Chapter of the Association for Computational Linguistics: Human Language Technologies}, pages 1304--1313. Association for Computational Linguistics (ACL).

\bibitem[{Chen et~al.(2017)Chen, Fisch, Weston, and Bordes}]{chen2017reading}
Danqi Chen, Adam Fisch, Jason Weston, and Antoine Bordes. 2017.
\newblock Reading wikipedia to answer open-domain questions.
\newblock In \emph{55th Annual Meeting of the Association for Computational Linguistics, ACL 2017}, pages 1870--1879. Association for Computational Linguistics (ACL).

\bibitem[{Cole et~al.(2023)Cole, Zhang, Gillick, Eisenschlos, Dhingra, and Eisenstein}]{cole2023selectively}
Jeremy~R Cole, Michael~JQ Zhang, Daniel Gillick, Julian~Martin Eisenschlos, Bhuwan Dhingra, and Jacob Eisenstein. 2023.
\newblock Selectively answering ambiguous questions.
\newblock \emph{arXiv preprint arXiv:2305.14613}.

\bibitem[{El-Yaniv and Wiener(2010)}]{ElYaniv2010OnTF}
Ran El-Yaniv and Yair Wiener. 2010.
\newblock \href {https://api.semanticscholar.org/CorpusID:10773394} {On the foundations of noise-free selective classification}.
\newblock \emph{J. Mach. Learn. Res.}, 11:1605--1641.

\bibitem[{He et~al.(2021)He, Liu, Gao, and Chen}]{he2021deberta}
Pengcheng He, Xiaodong Liu, Jianfeng Gao, and Weizhu Chen. 2021.
\newblock \href {https://openreview.net/forum?id=XPZIaotutsD} {Deberta: Decoding-enhanced bert with disentangled attention}.
\newblock In \emph{International Conference on Learning Representations}.

\bibitem[{Kadavath et~al.(2022)Kadavath, Conerly, Askell, Henighan, Drain, Perez, Schiefer, Hatfield-Dodds, DasSarma, Tran-Johnson et~al.}]{kadavath2022language}
Saurav Kadavath, Tom Conerly, Amanda Askell, Tom Henighan, Dawn Drain, Ethan Perez, Nicholas Schiefer, Zac Hatfield-Dodds, Nova DasSarma, Eli Tran-Johnson, et~al. 2022.
\newblock Language models (mostly) know what they know.
\newblock \emph{arXiv preprint arXiv:2207.05221}.

\bibitem[{Kamath et~al.(2020)Kamath, Jia, and Liang}]{kamath2020selective}
Amita Kamath, Robin Jia, and Percy Liang. 2020.
\newblock Selective question answering under domain shift.
\newblock \emph{arXiv preprint arXiv:2006.09462}.

\bibitem[{Kuhn et~al.(2023)Kuhn, Gal, and Farquhar}]{kuhn2023semantic}
Lorenz Kuhn, Yarin Gal, and Sebastian Farquhar. 2023.
\newblock Semantic uncertainty: Linguistic invariances for uncertainty estimation in natural language generation.
\newblock \emph{arXiv preprint arXiv:2302.09664}.

\bibitem[{Kwiatkowski et~al.(2019)Kwiatkowski, Palomaki, Redfield, Collins, Parikh, Alberti, Epstein, Polosukhin, Devlin, Lee et~al.}]{kwiatkowski2019natural}
Tom Kwiatkowski, Jennimaria Palomaki, Olivia Redfield, Michael Collins, Ankur Parikh, Chris Alberti, Danielle Epstein, Illia Polosukhin, Jacob Devlin, Kenton Lee, et~al. 2019.
\newblock Natural questions: a benchmark for question answering research.
\newblock \emph{Transactions of the Association for Computational Linguistics}, 7:453--466.

\bibitem[{Lin et~al.(2022)Lin, Hilton, and Evans}]{lin2022teaching}
Stephanie Lin, Jacob Hilton, and Owain Evans. 2022.
\newblock Teaching models to express their uncertainty in words.
\newblock \emph{arXiv preprint arXiv:2205.14334}.

\bibitem[{Liu et~al.(2023)Liu, Wu, Michael, Suhr, West, Koller, Swayamdipta, Smith, and Choi}]{liu2022were}
Alisa Liu, Zhaofeng Wu, Julian Michael, Alane Suhr, Peter West, Alexander Koller, Swabha Swayamdipta, Noah~A Smith, and Yejin Choi. 2023.
\newblock We're afraid language models aren't modeling ambiguity.
\newblock \emph{arXiv preprint arXiv:2304.14399}.

\bibitem[{Lopes et~al.(2020)Lopes, Farajian, Bawden, Zhang, and Martins}]{lopes2020document}
Ant{\'o}nio~V Lopes, M~Amin Farajian, Rachel Bawden, Michael Zhang, and Andr{\'e}~FT Martins. 2020.
\newblock Document-level neural mt: A systematic comparison.
\newblock In \emph{22nd Annual Conference of the European Association for Machine Translation}, pages 225--234.

\bibitem[{Maruf et~al.(2019)Maruf, Martins, and Haffari}]{maruf2019selective}
Sameen Maruf, Andr{\'e}~FT Martins, and Gholamreza Haffari. 2019.
\newblock Selective attention for context-aware neural machine translation.
\newblock \emph{arXiv preprint arXiv:1903.08788}.

\bibitem[{Min et~al.(2020)Min, Michael, Hajishirzi, and Zettlemoyer}]{min2020ambigqa}
Sewon Min, Julian Michael, Hannaneh Hajishirzi, and Luke Zettlemoyer. 2020.
\newblock Ambigqa: Answering ambiguous open-domain questions.
\newblock \emph{arXiv preprint arXiv:2004.10645}.

\bibitem[{OpenAI(2022)}]{chatgpt}
OpenAI. 2022.
\newblock \href {https://openai.com/blog/chatgpt} {Introducing chatgpt.}

\bibitem[{Parrish et~al.(2021)Parrish, Chen, Nangia, Padmakumar, Phang, Thompson, Htut, and Bowman}]{parrish2021bbq}
Alicia Parrish, Angelica Chen, Nikita Nangia, Vishakh Padmakumar, Jason Phang, Jana Thompson, Phu~Mon Htut, and Samuel~R Bowman. 2021.
\newblock Bbq: A hand-built bias benchmark for question answering.
\newblock \emph{arXiv preprint arXiv:2110.08193}.

\bibitem[{Pavlick and Kwiatkowski(2019)}]{pavlick2019inherent}
Ellie Pavlick and Tom Kwiatkowski. 2019.
\newblock Inherent disagreements in human textual inferences.
\newblock \emph{Transactions of the Association for Computational Linguistics}, 7:677--694.

\bibitem[{Piantadosi et~al.(2012)Piantadosi, Tily, and Gibson}]{piantadosi2012communicative}
Steven~T Piantadosi, Harry Tily, and Edward Gibson. 2012.
\newblock The communicative function of ambiguity in language.
\newblock \emph{Cognition}, 122(3):280--291.

\bibitem[{Pilault et~al.(2023)Pilault, Garcia, Bra{\v{z}}inskas, and Firat}]{pilault2023interactive}
Jonathan Pilault, Xavier Garcia, Arthur Bra{\v{z}}inskas, and Orhan Firat. 2023.
\newblock Interactive-chain-prompting: Ambiguity resolution for crosslingual conditional generation with interaction.
\newblock \emph{arXiv preprint arXiv:2301.10309}.

\bibitem[{Press et~al.(2022)Press, Zhang, Min, Schmidt, Smith, and Lewis}]{press2022measuring}
Ofir Press, Muru Zhang, Sewon Min, Ludwig Schmidt, Noah~A Smith, and Mike Lewis. 2022.
\newblock Measuring and narrowing the compositionality gap in language models.
\newblock \emph{arXiv preprint arXiv:2210.03350}.

\bibitem[{Pyatkin et~al.(2022)Pyatkin, Hwang, Srikumar, Lu, Jiang, Choi, and Bhagavatula}]{pyatkin2022reinforced}
Valentina Pyatkin, Jena~D Hwang, Vivek Srikumar, Ximing Lu, Liwei Jiang, Yejin Choi, and Chandra Bhagavatula. 2022.
\newblock Reinforced clarification question generation with defeasibility rewards for disambiguating social and moral situations.
\newblock \emph{arXiv preprint arXiv:2212.10409}.

\bibitem[{Rajpurkar et~al.(2016)Rajpurkar, Zhang, Lopyrev, and Liang}]{Rajpurkar2016SQuAD1Q}
Pranav Rajpurkar, Jian Zhang, Konstantin Lopyrev, and Percy Liang. 2016.
\newblock \href {https://api.semanticscholar.org/CorpusID:11816014} {Squad: 100,000+ questions for machine comprehension of text}.
\newblock In \emph{Conference on Empirical Methods in Natural Language Processing}.

\bibitem[{Rao and Daum{\'e}~III(2018)}]{rao2018learning}
Sudha Rao and Hal Daum{\'e}~III. 2018.
\newblock Learning to ask good questions: Ranking clarification questions using neural expected value of perfect information.
\newblock \emph{arXiv preprint arXiv:1805.04655}.

\bibitem[{Shridhar et~al.(2023)Shridhar, Stolfo, and Sachan}]{shridhar-etal-2023-distilling}
Kumar Shridhar, Alessandro Stolfo, and Mrinmaya Sachan. 2023.
\newblock \href {https://doi.org/10.18653/v1/2023.findings-acl.441} {Distilling reasoning capabilities into smaller language models}.
\newblock In \emph{Findings of the Association for Computational Linguistics: ACL 2023}, pages 7059--7073, Toronto, Canada. Association for Computational Linguistics.

\bibitem[{Tran et~al.(2022)Tran, Liu, Dusenberry, Phan, Collier, Ren, Han, Wang, Mariet, Hu et~al.}]{tran2022plex}
Dustin Tran, Jeremiah Liu, Michael~W Dusenberry, Du~Phan, Mark Collier, Jie Ren, Kehang Han, Zi~Wang, Zelda Mariet, Huiyi Hu, et~al. 2022.
\newblock Plex: Towards reliability using pretrained large model extensions.
\newblock \emph{arXiv preprint arXiv:2207.07411}.

\bibitem[{Voita et~al.(2019)Voita, Sennrich, and Titov}]{voita2019good}
Elena Voita, Rico Sennrich, and Ivan Titov. 2019.
\newblock When a good translation is wrong in context: Context-aware machine translation improves on deixis, ellipsis, and lexical cohesion.
\newblock In \emph{Proceedings of the 57th Annual Meeting of the Association for Computational Linguistics}, pages 1198--1212.

\bibitem[{Williams et~al.(2018)Williams, Nangia, and Bowman}]{multinli}
Adina Williams, Nikita Nangia, and Samuel Bowman. 2018.
\newblock \href {http://aclweb.org/anthology/N18-1101} {A broad-coverage challenge corpus for sentence understanding through inference}.
\newblock In \emph{Proceedings of the 2018 Conference of the North American Chapter of the Association for Computational Linguistics: Human Language Technologies, Volume 1 (Long Papers)}, pages 1112--1122. Association for Computational Linguistics.

\bibitem[{Yin et~al.(2021)Yin, Fernandes, Martins, and Neubig}]{yin2021does}
Kayo Yin, Patrick Fernandes, Andr{\'e}~FT Martins, and Graham Neubig. 2021.
\newblock When does translation require context? a data-driven, multilingual exploration.
\newblock \emph{arXiv preprint arXiv:2109.07446}.

\bibitem[{Yu et~al.(2019)Yu, Chen, Wang, Artzi, and Lei}]{Yu2019InteractiveCB}
L.~Yu, Howard Chen, Sida Wang, Yoav Artzi, and Tao Lei. 2019.
\newblock \href {https://api.semanticscholar.org/CorpusID:207852454} {Interactive classification by asking informative questions}.
\newblock \emph{ArXiv}, abs/1911.03598.

\bibitem[{Yuan et~al.(2023)Yuan, Malaviya, and Yatskar}]{yuan2023ambicoref}
Yuewei Yuan, Chaitanya Malaviya, and Mark Yatskar. 2023.
\newblock Ambicoref: Evaluating human and model sensitivity to ambiguous coreference.
\newblock \emph{arXiv preprint arXiv:2302.00762}.

\end{thebibliography}
\appendix

\section{Data Details}\label{app:data_details}
% \ec{maybe we should put an example of each in the appendix?}
% \mz{TODO: Add Dataset examples to appendix}
In Table~\ref{tab:dataset_raw_examples} include of raw examples from each of our MT, QA, and NLI datasets and in Table~\ref{tab:dataset_details} we incude dataset statistics. In matching interpretations from NaturalQuestions to AmbigQA disambiguations, we eliminate all examples where the NaturalQuestions answers do not appear in any of the AmbigQA interpretations and when it matches more than one interpretation.

% \mz{Include Table of dataset sizes and examples. Include disambiguated versions of the input.}

\section{Modeling Details}\label{app:model_details}
\subsection{Oracle Question Generation System}
We present our oracle clarification generation prompts for QA, NLI, and MT in Table~\ref{tab:qa_followup_gen_prompt}, Table~\ref{tab:nli_followup_gen_prompt}, and Table~\ref{tab:mt_followup_gen_prompt}, respectively. We do not provide GPT-3.5 any system prompt, and the entire body of the prompt is provided in a single user-side message. Note that for our MT oracle prompt, there is the risk of answer leakage, since the output translation is included in the prompt. However, we do not find this is an issue, as the generated followup questions and answers are always in the source language only.

\subsection{Experimental Setups and Additional Results}
\paragraph{Prompts}
We present the prompts for responding with clarification, without clarification, and for SelfAsk in Tables~\ref{tab:with_clarification},~\ref{tab:no_clarification}, and~\ref{tab:selfask}. These tables also demonstrate the variations in prompt between tasks, particularly in the instructions. We base our NLI instruction and class-to-token mapping on the prompts from~\cite{liu2022were}.

To perform our experiments with disambiguated inputs for QA and NLI, we use the same prompt as responding without clarificaiton, substituting the input with the disambiguate form of the input. For MT where disambiguations are given as additional context sentences, we simply prepend ``Context: \dots'' onto each user input, filling in the context sentence.

For sampling unambiguous examples for SelfAsk, we use the unambiguous examples labeled in the MT and QA datasets. For NLI, where all examples are labeled as ambiguous, we use examples where all 9 annotators interpreted the input the same way as unambiguous examples, as these demonstrate the least variation in user intents.

% \pargraph{When To Ask Results}
% In Table~\ref{tab:when_to_ask_results_full}, we present our full table of results on our determining when to ask for clarification subtask. Note that we only inlude instances where sytems improved by at least 10\% with clarification, as noticeable gains are a prerequisite to this task.

%%%%%%%%%%%%%%%%%%%%%%%%%%%%%%%%%%%%%%%%%%%%%%%%%%
% MT Followup Gen Prompt
%%%%%%%%%%%%%%%%%%%%%%%%%%%%%%%%%%%%%%%%%%%%%%%%%%
\begin{table*}
\centering
\footnotesize

\begin{tabular}{c p{4cm} p{8cm}}
\toprule
Task & Input ($x$) & Interpretations / Outputs ($y$)  \\ \midrule
\multirow{11.5}{*}{MT} & \multirow{5.5}{4cm}{That is so sweet!}
&  \textbf{Context:} You've been so wonderful to me these past couple of months.  \\
&& \textbf{Target:} C'est tellement adorable. \\ \cmidrule{3-3}
&&  \textbf{Context:} Try some - it's like a sugar explosion!  \\
&& \textbf{Target:} C'est tellement sucré. \\ \cmidrule{2-3}
& \multirow{5.5}{4cm}{I've never seen so much dough!}
&  \textbf{Context:} The pizza's in the oven, but there's still some dough left.  \\
&& \textbf{Target:} Je n'ai jamais vu autant de pâte ! \\ \cmidrule{3-3}
&&  \textbf{Context:} Here you are - you've earnt it.  \\
&& \textbf{Target:} Je n'ai jamais vu autant de thune ! \\ \midrule
\multirow{17}{*}{QA} & \multirow{6.5}{4cm}{When is episode 113 of dragon ball super coming out?}
&  \textbf{Disambig:} When is episode 113 of dragon ball super coming out for its original airdate?  \\
&& \textbf{Answer:} October 29, 2017 \\ \cmidrule{3-3}
&&  \textbf{Disambig:} When is episode 113 of dragon ball super coming out for its american airdate?  \\
&& \textbf{Answer:} June 1, 2019 \\ \cmidrule{2-3}
& \multirow{10}{4cm}{Who plays the science officer on star trek discovery?}
&  \textbf{Disambig:} Who plays the science officer on star trek discovery who is a chief engineer?  \\
&& \textbf{Answer:} Anthony Rapp \\ \cmidrule{3-3}
&&  \textbf{Disambig:} Who plays the science officer on star trek discovery who is a Kelpien?  \\
&& \textbf{Answer:} Doug Jones \\ \cmidrule{3-3}
&&  \textbf{Disambig:} Who plays science officer Michael Burnham on Star Trek Discovery?  \\
&& \textbf{Answer:} Sonequa Martin-Green \\ \midrule
\multirow{13.5}{*}{NLI} & \multirow{6.5}{4cm}{A large number of people were not willing to take the risk. / A small number of people were willing to take the risk.}
&  \textbf{Disambig:} A large number of people, but not all people, were not willing to take the risk.  \\
&& \textbf{Label:} entailment \\ \cmidrule{3-3}
&&  \textbf{Disambig:} A large number of people, and possibly all people, were not willing to take the risk.  \\
&& \textbf{Label:} neutral \\ \cmidrule{2-3}
& \multirow{6.5}{4cm}{We have not been able to find any scientific evidence that extraterrestrial life exists. / There is no scientific evidence that extraterrestrial life exists.}
&  \textbf{Disambig:} There is no scientific evidence to be found that extraterrestrial life exists.  \\
&& \textbf{Label:} neutral \\ \cmidrule{3-3}
&&  \textbf{Disambig:} There has been no scientific evidence collected that extraterrestrial life exists.  \\
&& \textbf{Label:} entailment \\
\bottomrule
\end{tabular}
% \vspace{-11pt}
\caption{Raw ambiguous examples from each dataset.}
\label{tab:dataset_raw_examples}
\end{table*}
%%%%%%%%%%%%%%%%%%%%%%%%%%%%%%%%%%%%%%%%%%%%%%%%%%

%%%%%%%%%%%%%%%%%%%%%%%%%%%%%%%%%%%%%%%%%%%%%%%%%%
% MT Followup Gen Prompt
%%%%%%%%%%%%%%%%%%%%%%%%%%%%%%%%%%%%%%%%%%%%%%%%%%
\begin{table*}
\centering
\footnotesize
\begin{tabular}{lrrrr}
\toprule
Task & Ambiguous $x$ &  Unambiguous $x$ & Sampled Interpretations $y_*$ & Total Interpretations $y_*$   \\ \midrule
NLI & 504 & 0 & 504 & 1008 \\
QA & 652 & 830 & 1482 & 2781 \\
MT & 88 & 176 & 352 & 352 \\
\bottomrule
\end{tabular}
\caption{Counts of ambiguous and unambiguous inputs for each task. We also include counts of the number of sampled interpretations used in our ``determining when to clarify'' evaluations and the total number of interpretations.}
\label{tab:dataset_details}
\vspace{-11pt}
\end{table*}
%%%%%%%%%%%%%%%%%%%%%%%%%%%%%%%%%%%%%%%%%%%%%%%%%%

%%%%%%%%%%%%%%%%%%%%%%%%%%%%%%%%%%%%%%%%%%%%%%%%%%
% QA Followup Gen Prompt
%%%%%%%%%%%%%%%%%%%%%%%%%%%%%%%%%%%%%%%%%%%%%%%%%%
\begin{table*}
\centering
\footnotesize
\begin{tabular}{p{15cm}}
\toprule
Given the Ambiguous Question and several possible Intended Interpretations, ask a Clarification Question and provide Clarification Responses corresponding to each Intended Interpretations. Here are two examples:\\
\\
Example 1:\\
Ambiguous Question: Who has the highest goals in world football?\\
Intended Interpretation 1: Who has the highest goals in men's world international football?\\
Intended Interpretation 2: Who has the highest goals all-time in men's football?\\
Intended Interpretation 3: Who has the highest goals in women's world international football?\\\\
Clarification Question: Are you referring to the highest goals in men's world international football, or the highest goals in women's world international football?\\
Clarification Response 1: The highest goals in men's world international football.\\
Clarification Response 2: The highest goals all-time in men's football.\\
Clarification Response 3: The highest goals in women's world international football.\\
\\
Example 2:\\
Ambiguous Question: Who won the last olympic men's hockey?\\
Intended Interpretation 1: Who won Olympic men's ice hockey in 2014?\\
Intended Interpretation 2: Who won Olympic men's ice hockey in 2010?\\
Intended Interpretation 3: Who won Olympic men's ice hockey in 2006?\\
Intended Interpretation 4: Who won the 2016 olympic men's field hockey?\\
Intended Interpretation 5: Who won the 2012 olympic men's field hockey?\\
Intended Interpretation 6: Who won the 2008 olympic men's field hockey?\\
Clarification Question: Which year? Are referring to field hockey or ice hockey?\\
Clarification Response 1: 2014, ice hockey.\\
Clarification Response 2: 2010, ice hockey.\\
Clarification Response 3: 2006, ice hockey.\\
Clarification Response 4: 2016, field hockey.\\
Clarification Response 5: 2012, field hockey.\\
Clarification Response 6: 2008, field hockey.\\
\\
Now do it yourself:\\
Ambiguous Question: \{\}\\
Intended Interpretation 1: \{\}\\
\dots\\
Intended Interpretation $k$: \{\}\\

\bottomrule
\end{tabular}
\caption{QA Followup Generation Prompt.}
\label{tab:qa_followup_gen_prompt}
\end{table*}
%%%%%%%%%%%%%%%%%%%%%%%%%%%%%%%%%%%%%%%%%%%%%%%%%%

%%%%%%%%%%%%%%%%%%%%%%%%%%%%%%%%%%%%%%%%%%%%%%%%%%
% NLI Followup Gen Prompt
%%%%%%%%%%%%%%%%%%%%%%%%%%%%%%%%%%%%%%%%%%%%%%%%%%
\begin{table*}
\centering
\footnotesize
\begin{tabular}{p{15cm}}
\toprule
Given the Ambiguous Phrase and two possible Intended Interpretations, ask a Clarification Question and provide two Clarification Responses corresponding to each Intended Interpretations. Here are two examples:\\
\\
Example 1:\\
Ambiguous Phrase: Jon will wash his car, and Mary will too.\\
Intended Interpretation 1: Jon will wash his car, and Mary will wash hers.\\
Intended Interpretation 2: Jon and Mary will both wash Jon's car.\\
Clarification Question: Will Jon and Mary wash the same or different cars?\\
Clarification Response 1: The same.\\
Clarification Response 2: Different.\\
\\
Example 2:\\
Ambiguous Phrase: The hospital is being sued by six foot doctors.\\
Intended Interpretation 1: The hospital is being sued by six podiatrists.\\
Intended Interpretation 2: The hospital is being sued by doctors who are six feet tall.\\
Clarification Question: Do you mean six podiatrists or doctors who are six feet tall.\\
Clarification Response 1: Podiatrists.\\
Clarification Response 2: Doctors who are six feet tall.\\
\\
Now do it yourself:\\
Ambiguous Phrase: \{\}\\
Intended Interpretation 1: \{\}\\
Intended Interpretation 2: \{\}\\
\bottomrule
\end{tabular}
\caption{NLI Followup Generation Prompt.}
\label{tab:nli_followup_gen_prompt}
\end{table*}
%%%%%%%%%%%%%%%%%%%%%%%%%%%%%%%%%%%%%%%%%%%%%%%%%%

%%%%%%%%%%%%%%%%%%%%%%%%%%%%%%%%%%%%%%%%%%%%%%%%%%
% MT Followup Gen Prompt
%%%%%%%%%%%%%%%%%%%%%%%%%%%%%%%%%%%%%%%%%%%%%%%%%%
\begin{table*}
\centering
\footnotesize
\begin{tabular}{p{15cm}}
\toprule
Given the Ambiguous Phrase and two possible Translations, ask a Clarification Question about the meaning of a specific word or phrase and provide two Clarification Responses corresponding to each Translation. Here are two examples:\\
\\
Example 1:\\
Ambiguous Phrase: Where are the bats?\\
Translation 1: Où sont les chauves-souris?\\
Translation 2: Où sont les battes?\\
Clarification Question: What type of bats do you mean?\\
Clarification Response 1: The small animals with wings.\\
Clarification Response 2: Sticks like you would use for a sport.\\
\\
Example 2:\\
Ambiguous Phrase: I love dates.\\
Translation 1: J'adore les rendez-vous.\\
Translation 2: J'adore les dattes.\\
Clarification Question: What type of dates are you referring to?\\
Clarification Response 1: Meetings or appointments with someone.\\
Clarification Response 2: The sweet fruit from the date palm tree.\\
\\
Now do it yourself:\\
Ambiguous Phrase: \{\}\\
Translation 1: \{\}\\
Translation 2: \{\}\\
\bottomrule
\end{tabular}
\caption{MT Followup Generation Prompt.}
\label{tab:mt_followup_gen_prompt}
\end{table*}
%%%%%%%%%%%%%%%%%%%%%%%%%%%%%%%%%%%%%%%%%%%%%%%%%%

%%%%%%%%%%%%%%%%%%%%%%%%%%%%%%%%%%%%%%%%%%%%%%%%%%
% Responding with Clarifications Prompt
%%%%%%%%%%%%%%%%%%%%%%%%%%%%%%%%%%%%%%%%%%%%%%%%%%
\begin{table*}
\centering
\footnotesize
\begin{tabular}{c p{12cm}}
\toprule
Role & Messages \\ \midrule
System &  Translate the English sentences into French. \\ \\
User & English: It's tight enough if you ask me. \\
Assistant &  Follow-Up Question: Are you referring to physical tightness or tension in a situation? \\
User & Follow-Up Response: I'm referring to a tense or stressful situation. \\
Assistant &  French: C'est déjà assez tendu à mon avis. \\
\vspace{-3pt} \\
\dots \\ \\
User & English: Why don't you care for them then? \\
Assistant &  Follow-Up Question: What do you mean by 'care for them'? \\
User & Follow-Up Response: I mean, why don't you take responsibility for them or look after them? \\
Assistant &  French: \textcolor{teal}{Alors pourquoi tu ne t'en occupes pas ?} \\
\bottomrule
\end{tabular}
\caption{Prompt for few-shot MT with clarification questions. The colored text represents the output generated from the LLM assistant.}
\label{tab:with_clarification}
\end{table*}
%%%%%%%%%%%%%%%%%%%%%%%%%%%%%%%%%%%%%%%%%%%%%%%%%%

%%%%%%%%%%%%%%%%%%%%%%%%%%%%%%%%%%%%%%%%%%%%%%%%%%
% Responding without Clarifications Prompt
%%%%%%%%%%%%%%%%%%%%%%%%%%%%%%%%%%%%%%%%%%%%%%%%%%
\begin{table*}
\centering
\footnotesize
\begin{tabular}{c p{12cm}}
\toprule
Role & Messages \\ \midrule
System &  Answer the question. \\ \\
User & Question: How many medals did australia win in the 2000 olympics? \\
Assistant &  Is a Follow-Up Question Needed Here? Yes. \\
          & Follow-Up Question: Are you asking about the number of gold, silver, bronze, or total medals that Australia won in the 2000 Olympics? \\
User &  Follow-Up Answer: The total number of medals that Australia won in the 2000 Olympics. \\
Assistant & Answer: 58. \\  \\ 
User & Question: Who plays erica on the last man on earth? \\
Assistant & Is a Follow-Up Question Needed Here? No. \\
          & Answer: Cleopatra Coleman. \\
\vspace{-3pt} \\
\dots \\ \\
User & Question: How long do contestants get to answer on jeopardy? \\
Assistant & Is a Follow-Up Question Needed Here? \textcolor{red}{No.} \\
\bottomrule
\end{tabular}
\caption{Prompt for few-shot QA with SelfAsk. The colored text represents the output that is scored by LLM assistant, and used as an uncertainty estimate. Sampled exemplars are always half unambiguous examples, not requiring follow-up questions, and half ambiguous examples, requiring follow-ups.}
\label{tab:selfask}
\end{table*}
%%%%%%%%%%%%%%%%%%%%%%%%%%%%%%%%%%%%%%%%%%%%%%%%%%

%%%%%%%%%%%%%%%%%%%%%%%%%%%%%%%%%%%%%%%%%%%%%%%%%%
% SelfAsk Prompt
%%%%%%%%%%%%%%%%%%%%%%%%%%%%%%%%%%%%%%%%%%%%%%%%%%
\begin{table*}
\centering
\footnotesize
\begin{tabular}{c p{12cm}}
\toprule
Role & Messages \\ \midrule
System &  For each Context, determine whether the Claim is True, False, or Inconclusive. \\ \\
User & Context: The professor admitted that no students wrote course evaluations, which is surprising. \\
     & Claim: It is surprising that no students wrote course evaluations. \\
Assistant & Answer: True. \\
\vspace{-3pt} \\
\dots \\ \\
User & Context: Many birds are attracted to the island because of the abundance of food. \\
     & Claim: Many birds are attracted to the island because of the abundance of trees.  \\
Assistant & Answer: \textcolor{teal}{Inconclusive.} \\
\bottomrule
\end{tabular}
\caption{Few-shot NLI prompt without clarification. The colored text represents the output generated from the LLM assistant.}
\label{tab:no_clarification}
\end{table*}
%%%%%%%%%%%%%%%%%%%%%%%%%%%%%%%%%%%%%%%%%%%%%%%%%%

\end{document}